\documentclass[runningheads]{llncs}
 \usepackage{comment} 
\usepackage{subfigure}

\usepackage[export]{adjustbox}
\usepackage[T1]{fontenc}
\usepackage{xcolor}
\usepackage{graphicx}
\usepackage{tabularx}
\begin{document}
\title{Privacy-Preserving Statistical Data Generation: Application to Sepsis Detection\thanks{Supported by organization Instituto de Ingeniería del Conocimiento, Hospital Universitario Son Llátzer, the Fundación Instituto de Investigación Sanitaria Illes Balears (Spain) and BBforTAI (PID2021-127641OB-I00 MICINN/FEDER).}}
\titlerunning{Privacy-Preserving Statistical Data Generation}
%
\author{Eric Macias-Fassio\inst{1,2}\orcidID{0009-0004-8275-5294} \and
Aythami Morales\inst{2}\orcidID{0000-0002-7268-4785} \and
Cristina Pruenza\inst{1}\orcidID{0009-0002-8565-4507} \and Julian Fierrez\inst{2}\orcidID{0000-0002-6343-5656} }
\authorrunning{E. Macias-Fassio et al.}
%
\institute{Instituto de Ingeniería del Conocimiento, Madrid, Spain\\
mail{\{eric.macias, cristina.pruenza\}@iic.uam.es} \and
BiDA-Lab, Universidad Autónoma de Madrid, 28049, Madrid, Spain\\
\email{\{aythami.morales, julian.fierrez\}@uam.es}\\
}
\maketitle              
\begin{abstract}
The biomedical field is among the sectors most impacted by the increasing regulation of Artificial Intelligence (AI) and data protection legislation, given the sensitivity of patient information. However, the rise of synthetic data generation methods offers a promising opportunity for data-driven technologies. In this study, we propose a statistical approach for synthetic data generation applicable in classification problems. We assess the utility and privacy implications of synthetic data generated by Kernel Density Estimator and K-Nearest Neighbors sampling (KDE-KNN) within a real-world context, specifically focusing on its application in sepsis detection. The detection of sepsis is a critical challenge in clinical practice due to its rapid progression and potentially life-threatening consequences. Moreover, we emphasize the benefits of KDE-KNN compared to current synthetic data generation methodologies. Additionally, our study examines the effects of incorporating synthetic data into model training procedures. This investigation provides valuable insights into the effectiveness of synthetic data generation techniques in mitigating regulatory constraints within the biomedical field.

\end{abstract}
\section{Introduction}
The exponential growth of Artificial Intelligence (AI) has sparked a revolutionary wave across various sectors with its profound impact particularly evident in the biomedical field. AI's ability to analyze vast amounts of data quickly and accurately has transformed medical research, diagnosis, and treatment. In recent years, there has been significant progress in the application of machine learning (ML) and deep learning models for early disease diagnosis \cite{siddiq2022use}. These methodologies have exhibited substantial potential in identifying a diverse range of medical conditions, including cancer \cite{sharma2021systematic}, cardiovascular disease \cite{miao2020using}, and Parkinson's disease \cite{makarious2022multi}. Through sophisticated algorithms and analysis of extensive datasets, these models can potentially identify subtle patterns and markers indicative of these conditions at their early stages.

However, many governments are introducing strict regulations for personal data processing and AI applications such as new European Union AI act \footnote{\url{https://artificialintelligenceact.eu/}}, CCPA\footnote{\url{https://oag.ca.gov/privacy/ccpa}} (Unitated States), and LGPD\footnote{\url{https://iapp.org/resources/article/brazilian-data-protection-law-lgpd-english-translation/}} (Brazil), which enforces data protection measures. A significant development in the regulatory landscape of AI has occurred with the enactment of the AI Act within the European Union. This legislative framework is designed to oversee and govern the application of AI models. In the realm of biomedical research, cautious consideration must be exercised when employing patient data for the training of AI models. Patient data, characterized by its sensitive nature, is subject to stringent protection under data protection laws, necessitating the preservation of privacy.

A solution that can potentially overcome these limitations involves the generation of fully synthetic data (SD) as an alternative to real data. SD is artificial data generated by a trained model and built to replicate real data by taking into account its distribution (mean, variance) and structure (e.g. correlation between attributes) \cite{el2019synthetic}. The utilization of SD generation emerges as a versatile methodology in machine learning, extending its applications across two domains: augmenting datasets to enhance model training \cite{shafique2023breast,strelcenia2023improving} and safeguarding the privacy of sensitive information \cite{rankin2020reliability}. 
Henceforth, this study introduces a straightforward technique for SD generation and conducts a comparative evaluation against state-of-the-art methodologies in terms of both utility and privacy considerations. The evaluation of these methods is performed within the context of a real-world application, specifically the early diagnosis of sepsis.

In more detail, the main contributions of this work are the following: 

\begin{itemize}
    \item We propose KDE-KNN, a statistical method to generate synthetic data for training and evaluating supervised learning algorithms. 
    \item We evaluated the utility and privacy of the generated synthetic data using different supervised algorithms in the context of sepsis detection. Our findings demonstrate that KDE-KNN outperforms existing methods in generating synthetic tabular data for sepsis detection.
    \item We assessed the generalization capacity of KDE-KNN using two real databases with more than 2000 patients. Our results suggest that KDE-KNN has certain advantages in terms of generalization over other methods.
\end{itemize}

\section{Related works}
We have divided this section in two parts: (i) synthetic tabular data generation approaches in Healthcare, and (ii) machine learning models for predicting sepsis.

\subsection{Synthetic tabular data generation approaches in Healthcare}
Synthetic data (SD) was first proposed and defined by Rubin \cite{rubin1993statistical} and Little \cite{little1993statistical} in 1993 as datasets consisting of records of individual synthetic values instead of real values. Nowadays, the concept of SD has evolved to encompass artificial data generated by trained models, designed to emulate real data by faithfully capturing its distributional (such as mean and variance) and structural attributes (including correlations between attributes) \cite{el2019synthetic}.
SD generation stands out as a highly promising yet largely underexploited technology for fulfilling privacy-preserving laws. In the biomedical sector, synthetic data generation has been mainly investigated in medical imaging \cite{8363678},  electronic health records (EHR) free-text content \cite{8621223} and EHR tabular data \cite{yale2020generation}. 

This study focused on synthetic EHR tabular data generation, as it is the predominant type of data used to develop ML models to aid health care decision-making \cite{hernadez2023synthetic}. 

Tabular healthcare-related data stored in EHR contain vast and diverse amounts of patient-related data. Typically, each row in a healthcare tabular dataset represents a single data record containing descriptive patient details such as date of birth, gender, and demographic information, along with sensitive attributes primarily consisting of longitudinal data. This longitudinal data comprises a series of medical events occurring at various time points, encompassing diagnoses, laboratory test results, and prescription information \cite{chong2021privacy}.

In the healthcare context, numerous approaches to generating synthetic data can be found in the literature. Among these, one widely utilized algorithm is the Synthetic Minority Oversampling Technique (SMOTE) \cite{chawla2002smote}, representing a straightforward method for generating synthetic tabular data \cite{sinha2023dasmcc}. This algorithm operates by synthesizing new data through interpolation of the existing samples. 
Another statistical approach to generate synthetic data involves Kernel Density Estimation (KDE) based models. Our framework for synthetic data generation in the healthcare context relies on KDE, chosen for its non-parametric nature and demonstrated efficacy, particularly in small datasets, which are prevalent in the biomedical field \cite{plesovskaya2021empirical}.
Additional methodologies used for SD generation involves the utilization of generative models, which include generative adversarial networks (GANs) and diffusion models (DM).

Since their inception in 2014 \cite{goodfellow2014generative}, GANs have demonstrated exceptional capability in the production of synthetic image data \cite{alqahtani2021applications}. For this reason, the application of GANs to other  data types, such as tabular data, is a popular topic in the AI research community \cite{hazra2020synsiggan}. Some GAN-based synthetic tabular data generation approaches in Healthcare are ehrGAN \cite{che2017boosting}, medGAN \cite{choi2017generating}, GcGAN \cite{yang2019grouped}. However, owing to the difficulties associated with training these models, as well as constraints related to sample size, we opt not to evaluate such models in this study.

On the other hand, we have diffusion models (DM). DM represent another class of generative models which have been widely used in the computer vision field. Notably, recent advancements have led to the development of architectures tailored to exploit diffusion models for tabular data, such as TabDDPM \cite{kotelnikov2023tabddpm}, which has demonstrated significant potential and promising outcomes in this regard. For these reasons, in this study we have evaluated the performance between SMOTE, TabDDPM and KDE-based generative models.

\subsection{Machine learning models for predicting sepsis}
Sepsis is defined as life-threatening organ dysfunction caused by a dysregulated host response to infection \cite{singer2016third}. In 2017, approximately 20\% of all global deaths were attributed to sepsis \cite{rudd2020global}. Early diagnosis of sepsis is crucial in the clinical setting, as it could help to significantly improve patient outcomes \cite{weber2023roles}, but early and accurate sepsis detection is still a challenging clinical problem \cite{alanazi2023machine}. For this reason, several ML algorithms have been designed to predict sepsis using retrospective data \cite{electronics11091507,giannini2019machine,horng2017creating,islam2019prediction,kausch2021physiological,nemati2018interpretable}. To our current knowledge, existing algorithms for sepsis prediction operate within a defined temporal window, typically forecasting the likelihood of sepsis onset within a specific time lapse, such as the next 24 hours. In our study, we want to overcome temporal constraints by seeking to predict the presence or absence of sepsis without temporal limitations. Therefore, we frame the task of sepsis detection as a classification problem, with the aim of addressing the question: Will patient A develop sepsis in the future? Furthermore, we substantiate our findings through validation in an external cohort for robustness and generalizability.

\section{Materials Methods}

In this study we have used 2 databases: i) Mannheim database (MaDB) used for training our models and build the synthetic datasets; ii) Son Llàtzer hospital database (SLDB) used as external validating dataset to evaluate the trained models.

\subsection{Mannheim database}
We used the University Medical Centre Mannheim database of patients admitted to intensive care unit (ICU) \cite{schamoni2022ensembling}. This database contains a total of 1275 patients, 979 with non-sepsis and 296 with sepsis. Initially, the MaDB comprised 42 timelines of features and the diagnosis of sepsis at each time step. However, for comparative analysis with the SLDB, it was necessary to align the feature sets. Consequently, only 27 features were found to be common between both databases. Among this features we have the age of the patient and lab results (Table \ref{table:1}).

For our study we did not use temporal data, instead we set a cut-off value at 9 hours as we estimated that in this time period all clinical tests could be performed and laboratory results could be collected. If a test has been performed several times during this period, the last value is used. In this way we constructed a dataset where our predictor variables were collected in that time interval and the objective was to predict whether a patient will develop sepsis in the future (classification problem), regardless of the time lapse window. 

The Mannheim database (MaDB), contains temporal data that allow precise tracking of sepsis onset times for patients, as evidenced in Table \ref{table:2}. The notable variability in the timing of sepsis manifestation within this dataset underscores its inherent heterogeneity. However, we do not use this temporal information, because we treat the detection of sepsis as a classification problem, knowing that this is a more challenging problem to solve. The MaDB has been used to train and test models and generate synthetic data.

\begin{table}[t!]
\caption{Description of the 27 variables present in the databases.}
\centering
\begin{tabular}{|l|l|l|} 
 \hline
 ID & Feature & Description \\ 
 \hline
 1 & heart\_rate & Number of heartbeats per minute \\ 
\hline
2 & leukocytes & Cells of the immune system \\ 
\hline
3 & temperature	& Body temperature \\ \hline
4 & respiratory\_rate & Number of breaths a person takes per minute \\ \hline 
5 & bilirubin & Compound originating from heme catabolism \cite{vitek2021bilirubin}\\ \hline 
6 & blood\_urea\_nitrogen & Amount of urea nitrogen in the blood \\ 
\hline
7 & creatinine & The end product of creatine phosphate metabolism \cite{kashani2020creatinine} \\ \hline
8 & diastolic\_bp	& Blood pressure measurement \\ 
\hline
9 & fraction\_of\_inspired\_o2	& Fraction of oxygen present in the air that a person  inhales \\ 
\hline
10 & systolic\_bp	& Blood pressure measurement \\ 
\hline
11 & thrombocytes & Blood cells \\ \hline
12 & lactate & Metabolite of glucose \\ 
\hline
13 & bicarbonate & Electrolyte \cite{shrimanker2019electrolytes} \\ \hline
14 & c-reactive\_protein & Molecule secreted in response to inflammatory cytokines \cite{du2000function}  \\ \hline
15 & hemoglobin & Protein found in red blood cells   \\ 
\hline
16 & lymphocytes & Cells of the immune system \\ 
\hline
17 & sodium & Electrolyte \cite{shrimanker2019electrolytes} \\ 
\hline
18 & pancreatic\_lipase  & Enzyme \cite{lowe1997structure} \\ \hline
19 & procalcitonin & Peptide \\ 
\hline
20 & oxygen\_saturation & Percentage of hemoglobin bound to oxygen \cite{hafen2018oxygen} \\ 
\hline
21 & blood\_glucose & Concentration of glucose \\ 
\hline
22	& chloride & Electrolyte \cite{shrimanker2019electrolytes} \\ \hline
23	& calcium & Electrolyte \cite{shrimanker2019electrolytes} \\ \hline
24	& potassium	& Electrolyte \cite{shrimanker2019electrolytes} \\ \hline
25 & alanine\_transaminase & Enzyme \cite{sookoian2012alanine} \\ 
\hline
26 & aspartate\_transaminase & Enzyme \cite{sookoian2012alanine}  \\ 
\hline
27	& age	& Years \\ 
 \hline
\end{tabular}
\label{table:1}
\end{table}

\begin{table}
\caption{Main characteristics of the databases, including the number of features and patients, as well as the mean, minimum and maximum time of sepsis (in hours) onset and the service where the data were collected.}
\centering
\begin{tabular}{|c|c|c|c|c|c|c|} 
 \hline
 DB & Patients & Features & Mean (t) & Min (t) & Max (t) & Hospital service \\ 
  \hline
 MaDB & 979 non-sepsis/ 296 sepsis & 27 & 208.7 & 39.5 & 1385 & ICU \\ 
 SLDB & 1014 non-sepsis/ 1014 sepsis & 27 & 36 & 24 & 48 & ICU/emergency\\ 
 \hline
\end{tabular}
\label{table:2}
\end{table}

\subsection{Son Llátzer hospital database}
We used a database from Son Llàtzer Hospital of patients admitted to emergency and ICU. The SLDB contains 2028 patients in total, 1014 with non-sepsis and 1014 with sepsis. In this database, we also selected the 27 common features with the MaDB. However, within the SLDB, the precise mean sepsis onset time remains unknown. According to insights from the medical team, the mean sepsis onset time is estimated to range between 24 to 48 hours. 
We employed this database for external validation, acknowledging significant disparities in sepsis onset times compared to our primary dataset. Notably, there are substantial variations in data distribution between the two databases. Thus, we perceived this as an opportunity to assess the generalization capacity of our models across diverse demographic populations.

\subsection{Sepsis Prediction Models}
Predicting sepsis onset remains a critical challenge in clinical practice due to its rapid progression and potentially life-threatening consequences \cite{singer2016third}. Early detection and intervention are paramount for improving patient outcomes and reducing mortality rates associated with this severe condition. Consequently, our study undertook an evaluation of three distinct ML models and assessed their performance based on the Area Under the Curve (AUC) score.

\begin{itemize}
    \item Random Forest (RF) \cite{breiman2001random}: RF is a widely used machine learning algorithm that belongs to the ensemble learning family, characterized by the construction of multiple decision trees during training. For classification tasks, RF outputs the predicted class, which in the context of sepsis prediction signifies whether a patient is likely to develop sepsis or not. 
    \item Support Vector Machine (SVM) \cite{cortes1995support}: SVM is a machine learning algorithm used for classification tasks. Unlike traditional classifiers that aim to find a decision boundary that separates classes, SVM seeks to find the hyperplane that best divides the classes while maximizing the margin between them. In our experiments we have used two SVM changing the type of kernel: \textit{i)} SVM with a linear kernel; \textit{ii)} SVM with a radial basis function (rbf) kernel.
    \item The hyperparameters of the models were tuned using the Optuna library \cite{optuna_2019}. Specifically, we employed a TPE (Tree-structured Parzen Estimator) sampler with 40 trials to maximize AUC. For the Random Forest (RF) model, the optimal hyperparameters were determined as follows: bootstrap was set to False, max\_depth to 20, max\_features to 5, min\_samples\_leaf to 5, min\_samples\_split to 12, and n\_estimators to 500. As for the SVM models, both of them had C set to 1. Other parameters that were not mentioned stay by default according to scikit-learn library \cite{scikit-learn}.
\end{itemize}



\subsection{Statistical Data Modelling Approaches}

In this paper we have analysed 3 popular statistical data modelling methods:

\textbf{SMOTE} \cite{chawla2002smote}: SMOTE is an oversampling methodology that was initially employed to generate synthetic observations exclusively from the minority class. However, we expanded this approach to incorporate the majority class as well \cite{kotelnikov2023tabddpm}, resulting in the creation of a fully synthetic dataset.

\textbf{TabDDPM} \cite{kotelnikov2023tabddpm}: TabDDPM is a design of denoising diffusion probabilistic models for tabular data. To tackle mixed-type characteristics of tabular data, this architecture integrates gaussian diffusion for capturing the characteristics of continuous features and multinomial diffusion for effectively modeling categorical attributes.

\textbf{KDE} \cite{parzen1962estimation,Rosenblatt1956RemarksOS}: Kernel Density Estimation (KDE) was proposed by Rosenblatt \cite{Rosenblatt1956RemarksOS} and Parzen \cite{parzen1962estimation}. KDE is a method used to estimate probability density functions. By constructing this distribution, we gain the ability to generate synthetic data samples through sampling. This capability allows for the creation of synthetic datasets representative of the underlying probability distribution. We conducted experiments using the multivariate KDE. In the multivariate KDE technique, we constructed distributions jointly, taking into account the interdependencies between features. This allowed us to capture complex relationships and dependencies across multiple variables simultaneously.

\subsection{KDE-KNN: Privacy Preserving Synthetic Clinical Data}

Our proposed methodology is founded on the integration of KDE and K-Nearest Neighbors (KNN) model. The idea is to use a multivariate gaussian KDE to approximate the probability density function of the original dataset features and then sample from it to generate synthetic datasets. However, as the feature space can be very large, we train a KNN to validate the synthetic samples. The procedural steps undertaken to construct our synthetic dataset are the following:

\begin{enumerate}
    \item Training a KNN model: A K-Nearest Neighbors (KNN) model was trained using the provided training dataset.
    \item Data preparation for KDE: The training dataset was partitioned into two distinct groups: patients with sepsis ($18.55\%$) and patients without sepsis ($81.45\%$).
    \item Multivariate KDE construction: Statistical independent multivariate KDE distributions were trained for each subgroup.
    \item Sampling synthetic data: Sampling was performed from each multivariate KDE model, generating 540 synthetic patients with sepsis and 540 synthetic patients without sepsis.
    \item Validation using KNN model: Validation of the synthetic samples was conducted by the trained KNN model. Synthetic data originated from the KDE model built with non-sepsis data should be classified as non-sepsis by the KNN model. Any discrepancies lead to the discarding of the data point.
\end{enumerate}
 
 Steps 4 and 5 were iteratively executed until we attained a total of 540 synthetic data points for patients with sepsis and 540 synthetic data points for patients without sepsis. This process ensures the creation of a balanced synthetic dataset representative of both septic and non-septic patient populations.

\begin{figure}[t!]
\centering
\includegraphics[width=\textwidth]{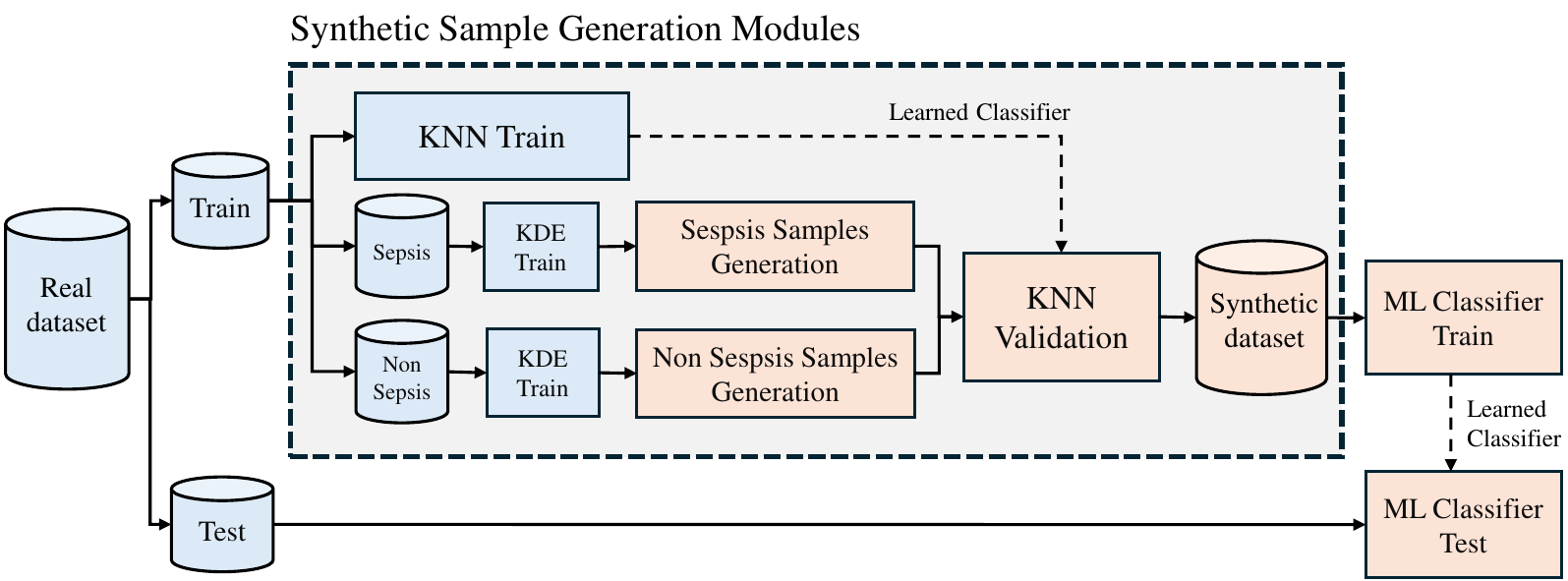}
\caption{KNN-KDE synthetic method block diagram including the generation modules based on two Kernel Density Estimators (Sepsis and Non-Sepsis) and K-NN sampling. } 
\label{fig3:3}
\end{figure}

For clarification, we closed this section by visualizing our proposed synthetic method as a flowchart, illustrated in Fig. \ref{fig3:3}. 

\section{Experiments and Results}
In this section, we assess the influence of synthetic data on the performance of sepsis detection models.
\subsection{Experimental Protocol}
Initially, our study is based on two distinct sepsis databases: the MaDB and the SLDB. The MaDB served as the primary dataset for model training/testing and synthetic data generation, while the SLDB was exclusively utilized for external validation purposes. 

Our first experimental phase involved evaluating model performance using real data exclusively. To accomplish this, we employed the MaDB and we partitioned the data into training (85\%) and testing (15\%) sets, repeating the experiment three times while changing the seed. Additionally, each partition underwent an external validation using the SLDB. Notably, our analysis revealed that the performance of the most effective models remained consistent across different partitions, suggesting minimal impact of partitioning on model performance. 

The second phase of experimentation was dedicated to assessing the utility of synthetic data. Our focus was on evaluating model performance exclusively using synthetic data. To achieve this, we generated a fully synthetic balanced dataset comprising 540 samples with sepsis and an equal number of 540 samples without sepsis. This balanced dataset mirrored the size of our original imbalanced training set. Building upon the stability observed in model performance across the three partitions in Experiment 1, one of the partitions was randomly selected for this subsequent analysis. This selection yielded both a training set and a test set sourced from the MaDB. The training set acted as a seed for generating synthetic data, employing the methodologies outlined in Section 3.4. For each method, three distinct synthetic balanced datasets were generated.
Following this, the quality of the synthetic data was evaluated using the test set. Additionally, an external validation was conducted using the SLDB to ensure the reliability and validity of our findings. The results of this experiment showed that the best method to generate synthetic data in the context of sepsis prediction is KDE-KNN. Finally, in the experiment 3, we examined how the incorporation of both real and synthetic data in the training set influenced model performance.

The third phase of our experimentation aimed to assess data privacy preservation using KDE-KNN method. In the experiment 4, we investigated the proximity of synthetic data to real data using the Mean Distance to Closest Record (DCR) metric \cite{zhao2021ctab}. Mean DCR calculates the average distance between synthetic samples and their closest real data points.

\subsection{Real Data and Synthetic Data Utility}
The findings from Experiment 1 are presented in Table \ref{tab3:3}. This experiment involved evaluating model performance in terms of AUC using exclusively real data. The results indicated that the RF model demonstrated superior performance in the test set. Nevertheless, concerning generalization, the results indicated a
a lower performance of the RF model, while the SVM with the rbf kernel demonstrated better generalization capabilities.

\begin{table}[t!]
\caption{Results of Experiment 1 using real data. The result is shown in terms of AUC $\pm$ variance as each model was trained 3 times.  }\label{tab3:3}
\centering
\begin{tabular}{|l|c|c|}
\hline
Model &  MaDB & SLDB\\
\hline
RF &  {\bfseries $0.6708 \pm 0.0169$} & $0.6469 \pm 0.0313$\\
SVM lineal kernel & $0.5426 \pm 0.0581$ & $0.6120 \pm 0.0701$\\
SVM rbf kernel & $0.6194 \pm 0.0119$ & \bfseries{$0.6952 \pm 0.0282$} \\
\hline
\end{tabular}
\end{table}

Experiment 2 aimed to assess the utility of synthetic data. The outcomes of Experiment 2 are detailed in Table \ref{tab4:4}. This experiment involved evaluating model performance using balanced synthetic datasets. 

The findings indicated a notable enhancement in model performance when employing balanced synthetic datasets. Remarkably, balanced synthetic data appeared to outperform real imbalanced data. Specifically, the SVM model with the rbf kernel demonstrated superior performance when trained on synthetic data generated using the KDE-KNN method. Furthermore, enhanced model performance was evident in the external validation database. These results may be due to the reduced heterogeneity of the external database and the earlier onset of sepsis in patients, suggesting that our models perform better when sepsis occurs within the 24-48 hour timeframe. Additionally, we aim to emphasize the minimal variance observed in synthetic datasets generated through our method.

\begin{figure}[t!]
\centering
\includegraphics[width=0.9\textwidth]{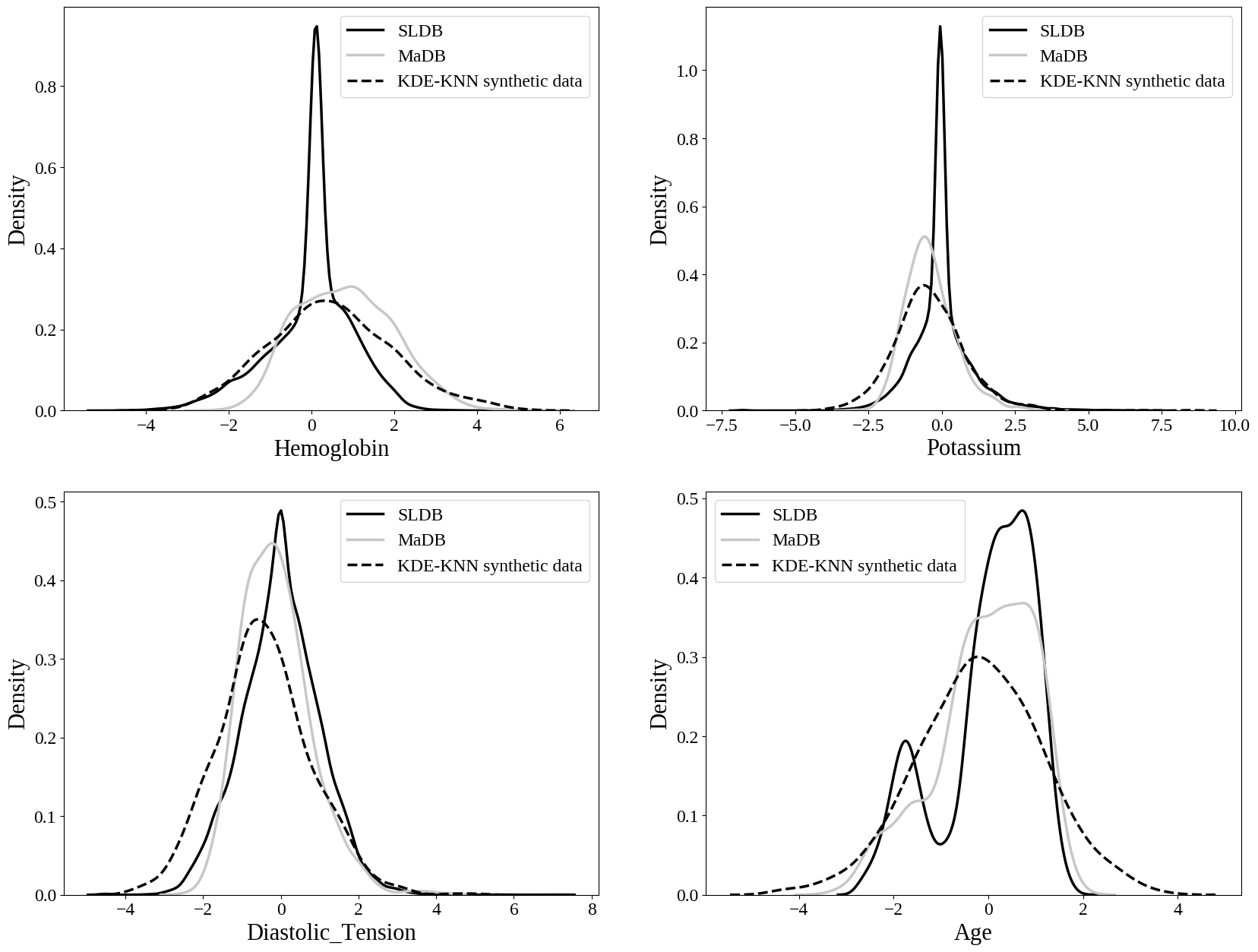}
\caption{Distribution of $4$ features from the two real datasets and the synthetic dataset. The solid black line represents the data distribution from the SLDB, the grey line represents the distribution from MaDB and the dashed line represents the distribution of synthetic data generated by KDE-KNN. All features were normalized using a z-score normalization technique.} 
\label{fig1:1}
\end{figure}

In Experiment 3, our goal was to examine how the combination of real and synthetic data during training affects model performance. The findings from Experiment 3 are presented in Table \ref{tab5:5}. For this analysis, we selected the best model and the best synthetic method from Experiment 2, which were identified as the SVM model with an rbf kernel and KDE-KNN as the synthetic method. We proceeded to train the SVM model using varying proportions of real and synthetic data generated by KDE-KNN, as illustrated in Table \ref{tab5:5}. Experiments combining real and synthetic data were performed 3 times using different seeds to sample the data. The findings indicated that augmenting the percentage of synthetic data generated with KDE-KNN in the training set leads to an improvement in the model performance, attributable to the enhanced balance of the dataset. The Fig. \ref{fig1:1} shows the normalized distributions of $4$ features in the real and synthetic databases. Note that the differences between both real databases are significant, and how the distribution of the synthetic samples tend to be realistic.   

\begin{table}[t!]
\caption{Results of experiment 2 using synthetic data. The result is shown in terms of AUC $\pm$ variance as each model was trained with 3 synthetic datasets.}\label{tab4:4}
\centering
\begin{tabular}{|l|l|c|c|}
\hline
Method & Model &  MaDB & SLDB\\
\hline
 & RF &  $0.6721 \pm 0.0144$ & $0.4560 \pm 0.0491$\\
SMOTE & SVM (lineal) & $0.6309 \pm 0.0346$ & $0.6583 \pm 0.0453$\\
 & SVM (rbf) & $0.6771 \pm 0.0212$ & $0.4437 \pm 0.0596$\\
\hline
& RF &   $0.6942 \pm 0.0102$ & $0.5187 \pm 0.0804$\\
TabDDPM \cite{kotelnikov2023tabddpm} & SVM (lineal) & $0.6697 \pm 0.0207$ & $0.6446 \pm 0.046$\\
 & SVM (rbf) & $0.7020 \pm 0.0095$ & $0.6949 \pm 0.0246$ \\
\hline
 & RF &  $0.6495 \pm 0.0051$ & $0.6261 \pm 0.00255$\\
KDE & SVM (lineal) & $0.6449 \pm 0.0017$ & $0.7202 \pm 0.0215$\\
 & SVM (rbf) & $0.6748 \pm 0.0072$ & $0.7114 \pm 0.0019$ \\
\hline
 & RF &  $0.6914 \pm 0.0097$ & $0.7650 \pm 0.0049$\\
\bfseries{KDE-KNN [ours]} & SVM (lineal) & $0.7092 \pm 0.0064$ & $0.7541 \pm 0.0040$\\
 & SVM (rbf) & \bfseries{$0.7129 \pm 0.0062$} & \bfseries{$0.7682 \pm 0.0016$} \\
\hline
\end{tabular}
\end{table}

\begin{table}
\caption{Results of experiment 3, combining real and synthetic data in the training set using the SVM model with rbf kernel. The results are shown in terms of AUC $\pm$ variance.}\label{tab5:5}
\centering
\begin{tabular}{|c|c|c|c|}
\hline
\% Real & \% Synthetic &  MaDB & SLDB\\
\hline
100 & 0 &  $0.6267 \pm 0$ & $0.6706 \pm 0$\\
80 & 20 &  $0.6828 \pm 0.0177$ & $0.7329 \pm 0.0121$\\
60 & 40 &  $0.6874 \pm 0.0047$ & $0.7319 \pm 0.0195$\\
40 & 60 &  $0.7033 \pm 0.0066$ & $0.7515 \pm 0.0090$\\
20 & 80 &  $0.7160 \pm 0.0099$ & $0.7589 \pm 0.0079$\\
0 & 100 &  $0.7129 \pm 0$ & $0.7682 \pm 0$\\
\hline
\end{tabular}
\end{table}

\subsection{Privacy preservation result}

\begin{figure}[t!]
\centering
\includegraphics[width=\textwidth]{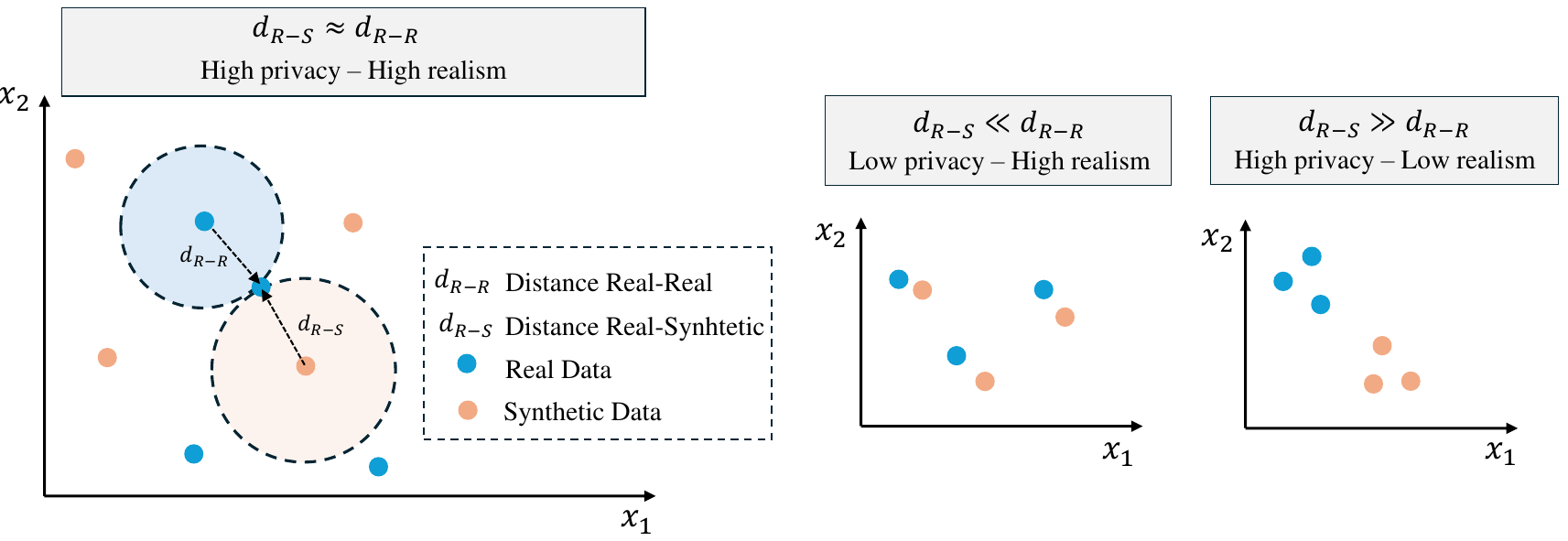}
\caption{Compromise between privacy and realism of synthetic samples. The graphs represent the distance between real and synthetic samples in a conceptual 2-dimensional space.} 
\label{fig: privacy-preserving}
\end{figure}

In Experiment 4, we conducted an analysis of the mean Distance to Closest Record (DCR) \cite{zhao2021ctab} between synthetic samples and their nearest real data points. The DCR is calculated as the Euclidean distance between a real sample and the closest synthetic sample. Low DCR values suggest that synthetic samples closely resemble real data points, potentially compromising privacy requirements. Conversely, higher DCR values indicate that the generative model can produce novel records rather than mere replicas of existing data. It is important to note that out-of-distribution data, such as random noise, can also yield high DCR values. Therefore, DCR must be evaluated alongside machine learning efficiency considerations \cite{kotelnikov2023tabddpm}. The Fig. \ref{fig: privacy-preserving} illustrates the compromise between privacy-preserving generation and realism of the synthetic samples.

\begin{figure}[t!]
\centering
\includegraphics[width=0.6\textwidth]{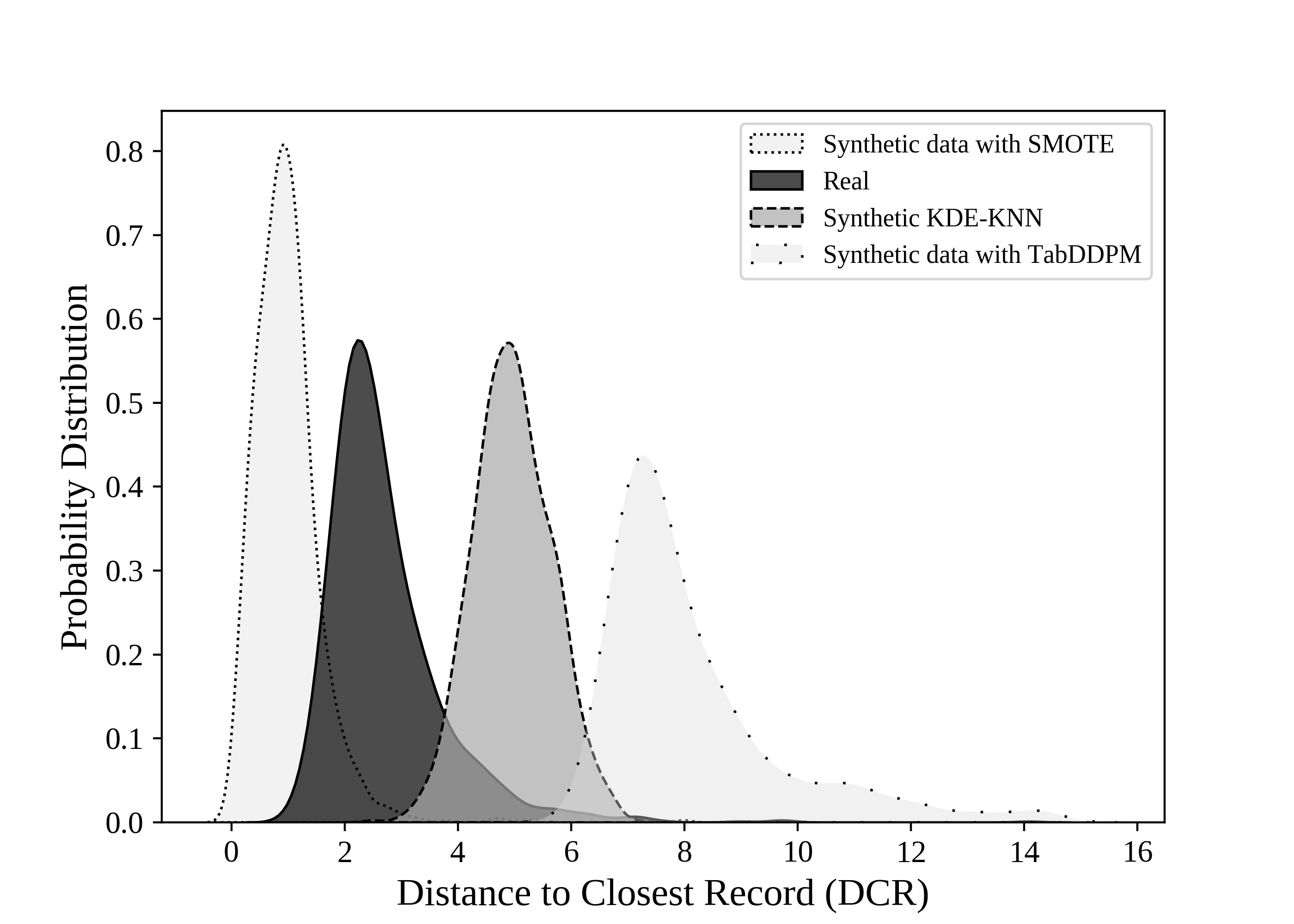}
\caption{Probability distribution of the Distance to Closest Record (DCR) for real samples and synthetic samples generated with the $3$ generation approaches evaluated in our experiments.} 
\label{fig: DCR}
\end{figure}

The Fig. \ref{fig: DCR} presents the probability distributions of DCR for the real samples ($d_{R-R}$) and the $3$ generation approaches evaluated in previous experiments ($d_{R-S}$). For SMOTE, the mean DCR value was 0.989, while for KDE-KNN and TabDDPM, the values were 4.971 and 7.463 respectively. Comparing these results with the mean distance between real data, which was 2.715, we observe that both TabDDPM and KDE-KNN demonstrate efficacy in generating synthetic data that preserves privacy, exhibiting superior performance compared to SMOTE.

\section{Conclusions}
Motivated by the imperative of adhering to data privacy regulations, we introduce KDE-KNN, a statistical method for generating tabular synthetic data. Through an extensive evaluation within the context of sepsis detection, we assessed this method in terms of both utility and privacy. Remarkably, our findings suggested that synthetic data outperformed real data in sepsis detection. We attributed this phenomenon to the fact that real dataset was quite imbalanced while synthetic dataset was balanced. For this reason,  KDE-KNN, also would be a good method to balance datasets. Moreover, our findings have been corroborated through validation in an external database, reinforcing the  generalizability potential of our synthesis approach.
Additionally, our results affirmed the efficacy of KDE-KNN in preserving privacy, as evidenced by the distance observed between synthetic and real data points. In conclusion, KDE-KNN emerges as a promising method for not only enhancing dataset utility but also safeguarding data privacy, making it a valuable tool in various data-driven applications.

\bibliographystyle{splncs04}
\bibliography{bibliography}
%
%
%
%

\end{document}